\documentclass[journal,12pt,onecolumn,draftclsnofoot,]{IEEEtran}
\usepackage[T1]{fontenc}
\usepackage{amsmath,amssymb}

\usepackage{todonotes}
\usepackage{latexsym}
\usepackage{algorithm}
\usepackage{algorithmic}
\usepackage{booktabs}
\usepackage{colortbl,etoolbox,siunitx,xcolor}
\usepackage{bm}
\usepackage{colortbl}
\usepackage{amsmath}
\usepackage{subcaption}
\usepackage{placeins}
\usepackage{listings}
\usepackage{color}
\definecolor{dkgreen}{rgb}{0,0.6,0}
\definecolor{gray}{rgb}{0.5,0.5,0.5}
\definecolor{mauve}{rgb}{0.58,0,0.82}
\robustify\bfseries \robustify\cellcolor \newcommand{\ud}{\,\mathrm{d}}
\newcommand{\ngbf}[1]{\mathbf{#1}}
\newcommand{\gbf}[1]{\boldsymbol{#1}}
\newcommand{\E}{\mathrm{E}}
\newcommand{\Cov}{\operatorname{Cov}}
\newcommand{\Var}{\operatorname{Var}}
\renewcommand{\E}{{\rm I\kern-.3em E}}

\DeclareMathOperator*{\argmin}{arg\,min}
\begin{document}
\title{Cluster-based Kriging Approximation Algorithms for Complexity Reduction}
%\author{\IEEEauthorblockN{Bas van Stein, Hao Wang, Wojtek Kowalczyk,  Michael Emmerich and Thomas B\"ack}
%\IEEEauthorblockA{Leiden Institute of Advanced Computer Science\\
%Leiden University\\
%Niels Bohrweg 1, Leiden, The Netherlands\\
%Email: \{b.van.stein, h.wang, w.j.kowalczyk, m.t.m.emmerich, t.h.w.baeck\}@liacs.leidenuniv.nl}
%}

\author{Bas van Stein,
        Hao Wang,
        Wojtek Kowalczyk,
        Michael Emmerich,\\
        and~Thomas B\"ack% <-this % stops a space
\thanks{Leiden Institute of Advanced Computer Science}
\thanks{Niels Bohrweg 1, Leiden, The Netherlands.}
\thanks{e-mail: b.van.stein@liacs.leidenuniv.nl}% <-this % stops a space
}

%\markboth{IEEE Computational Intelligence Magazine}%
%{van Stein \MakeLowercase{\textit{et al.}}: Cluster Based Kriging Approximation Algorithms}

\bibliographystyle{IEEEtran}

\maketitle

\begin{abstract}
\emph{Kriging} or \emph{Gaussian Process Regression} is applied in many fields as a non-linear regression model as well as a surrogate model in the field of evolutionary computation. However, the computational and space complexity of Kriging, that is cubic and quadratic in the number of data points respectively, becomes a major bottleneck with more and more data available nowadays. In this paper, we propose a general methodology for the complexity reduction, called cluster Kriging, where the whole data set is partitioned into smaller clusters and multiple Kriging models are built on top of them. In addition, four Kriging approximation algorithms are proposed as candidate algorithms within the new framework. Each of these algorithms can be applied to much larger data sets while maintaining the advantages and power of Kriging. The proposed algorithms are explained in detail and compared empirically against a broad set of existing state-of-the-art Kriging approximation methods on a well-defined testing framework.  According to the empirical study, the proposed algorithms  consistently outperform the existing algorithms. Moreover, some practical suggestions are provided for using the proposed algorithms.
\end{abstract}


\section{Introduction}
\label{sec:Intro}

%Introduction

\emph{Kriging}, or \emph{Gaussian Process Regression}~\cite{rasmussen2006gaussian} is a popular and elegant kernel based regression model capable of modeling very complex functions. Kriging is used in many fields e.g. engineering, mining and geology, as a tool for the analysis of data sets, for prediction purposes and for \emph{Surrogate model based optimization} \cite{simpson2001kriging}.
Many other regression models exist, such as parametric models, which are easy to interpret but may lack expressive power to model complex functions. 
On the other hand, \emph{Regression Tree} based methods like \emph{Random Forests} \cite{Breiman:2001:RF:570181.570182} or \emph{Gradient Boosted Decision Trees} lack the advantage of interpretation~\cite{d2012accurate} but have more expressive power.
Another method is \emph{Linear Model Trees} \cite{torgo1997functional}, which uses a tree structure with linear models at the leaves of the tree. There are also more complex algorithms like \emph{Neural Networks}, or \emph{Extreme Learning Machines} \cite{Huang2011}, that are able to model very complex functions but are usually not easy to work with in practice. There are also different kernel based methods such as \emph{Support Vector Machines} \cite{vapnik2013nature} and \emph{Radial Basis Functions} \cite{buhmann2004radial}. The main advantage of Kriging over other regression methods is that Kriging provides not only the estimate of the value of a function, but also the mean squared error of the estimation, the so-called \emph{Kriging variance}. The Kriging variance can be seen as the uncertainty assessment of the model and has been exploited in surrogate model based optimization and many other applications. 
Despite the clear advantage of the Kriging variance, Kriging suffers from one major problem, the high training time and space complexity, which are $O(n^3)$ and $O(n^2)$, respectively. Where $n$ denotes the number of points. To overcome this complexity problem, Kriging approximation algorithms such as \cite{Tresp2000} and \cite{Hartman2008} are introduced. Unfortunately, these approximation algorithms are usually less accurate than the original Kriging algorithm.

\textbf{Contributions}. An overview of Kriging approximation methods is presented and a novel divide and conquer based approach,  \emph{Cluster Kriging} (CK), is introduced. The novel Cluster Kriging framework contains three steps, Partitioning, Modeling and Predicting. Each of the steps can be implemented using a wide range of approaches, which are explained in detail in this paper. Using these approaches four algorithms are implemented and compared against each other and the state of the art. One particular interesting and novel algorithm that uses the Cluster Kriging methodology is the proposed \textbf{Model Tree Cluster Kriging} (MTCK). MTCK uses a regression tree with a specified number of leaf nodes to partition the data in the objective space. A Kriging model is then build on each partition defined by the tree's leaves. MTCK uses only one of the trained Kriging models per unseen record to predict, depending on which leaf node the unseen record is assigned to. The proposed algorithms are evaluated and compared to several state-of-the-art alternative Kriging approximation algorithms.  A well-defined testing framework for Kriging approximation algorithms \cite{Chalupka2012} is adopted for the comparisons.

\section{Kriging}
\label{sec:Kriging}
\textbf{Notation}. Throughout this paper, we shall use $n, k, d$ to denote the number of data points, the number of clusters and the dimensionality of the input space, respectively. In addition, the regression function is denoted as $f:\mathbb{R}^d \rightarrow \mathbb{R}$. In complexity statements in this paper we ignore $d$ since Kriging is generally used on low dimensional datasets. Without loss of generality, the column vector convention is adopted as the notation used throughout this paper.

%Kriging is a widely used non-parametric regression method that is able to model relations in data sets as nonlinear regression functions.
Loosely speaking, Kriging is a stochastic interpolation method in which the output value of a stochastic process is predicted as a linear function of the observed output values~\cite{stein1999interpolation,kleijnen2009kriging}. In particular, Kriging is the best linear unbiased predictor (BLUP) and the corresponding mean squared error of prediction is used for uncertainty qualification. Kriging originates from the field of spatial analysis/geostatistics and more recently is being widely used in Bayesian optimization and design and analysis of computer experiments (DACE)~\cite{jones1998efficient,sacks1989design}. The model features in providing the theoretical uncertainty measurement of estimations.

When the stochastic process is assumed to be Gaussian, Kriging is equivalent to Gaussian Process Regression (GPR), where the \emph{posterior} distribution of the regression function (posterior Gaussian process) is inferred through Bayesian statistics. In this paper, we shall consider this special case and adopt the mathematical treatment of the Gaussian process. Assume that input data points are summarized in the set $\mathcal{X}=\{{\ngbf{x}^{(1)}},\ngbf{x}^{(2)}, \ldots, \ngbf{x}^{(n)}\} \subseteq \mathbb{R}^d$ and the corresponding output variables are represented as $\mathbf{y}= [y(\ngbf{x}^{(1)}),y(\ngbf{x}^{(2)}),\ldots,y(\ngbf{x}^{(n)})]^\top$. Specifically, the mostly used variant of Kriging, \emph{Ordinary Kriging}, models the regression function  $f$ as a random process, that is a combination of an \emph{unknown} constant trend $\mu$ with a centered Gaussian Process $\varepsilon$. The output variables are considered as the ``noisy'' observation of $f$, that is perturbed by a Gaussian random noise $\gamma$:
\begin{align*}
	y(\ngbf{x}) &= f(\mathbf{x}) + \gamma(\ngbf{x}) = \mu + \varepsilon(\ngbf{x}) + \gamma(\ngbf{x}),
\end{align*}
$$\varepsilon(\mathbf{x}) \sim \mathcal{N}(0, \sigma_{\varepsilon}^2(\mathbf{x})), \quad \gamma(\ngbf{x}) \sim \mathcal{N}(0, \sigma_{\gamma}^2), \quad \varepsilon \text{ and } \gamma \text{ are independent}$$
Note that the noises (error terms) $\gamma$ are assumed to be homoscedastic (identically distributed) and independent, both from each other and the Gaussian Process $\varepsilon$. The centered Gaussian process $\varepsilon$ is a stochastic process which possesses zero mean everywhere and any finite collection of its random variables has a joint Gaussian distribution~\cite{rasmussen2006gaussian}.  It can be completely specified by providing a covariance function $k(\cdot,\cdot)$ to calculate the pairwise covariance:
\begin{equation*}
	 \Cov[\varepsilon(\ngbf{x}), \varepsilon(\ngbf{x}^{\prime})]= \E[\varepsilon(\ngbf{x})\cdot \varepsilon(\ngbf{x}^{\prime})]=k(\ngbf{x}, \ngbf{x}^{\prime}).
\end{equation*}
The covariance function $k(\cdot,\cdot)$ is a kernel function performing the so-called ``kernel trick'', which computes the inner product on the feature space as a function in the input space.
In this paper, the covariance function is chosen to be \emph{stationary}, meaning that the kernel $k$ is a function of $\mathbf{x}-\mathbf{x}^{\prime}$ and invariant to translations in the input space. Consequently, the variance $\sigma_{\varepsilon}^2(\mathbf{x})$ of a Gaussian process $\varepsilon$ is independent from the input $\mathbf{x}$ and thus denoted as $\sigma_{\varepsilon}^2$ in the following.
In practice, a common choice is the Gaussian covariance function (also known as squared exponential kernel):
\begin{equation}
k(\ngbf{x}, \ngbf{x}^{\prime}) = \sigma_{\varepsilon}^2\prod_{i=1}^{d}\exp\left(-\theta_i(x_i - x_i^{\prime})^2\right), \label{eq:gaussian-kernel}
\end{equation}
where $\theta_i$'s are called hyper-parameters, that are either predetermined or estimated through model fitting, and $\sigma_{\varepsilon}^2$ is inferred by maximum likelihood method. We omit the likelihood function in this paper. To infer output value $y^{(t)} = y(\mathbf{x}^{(t)})$ at an unobserved data point $\mathbf{x}^{(t)}$, the joint distribution of $y^{(t)}$ and observed outputs $\mathbf{y}$ are derived, conditioning on the input data set $\mathcal{X}$, $\ngbf{x}^{(t)}$ and the unknown prior mean $\mu$. Such a joint distribution is a multivariate Gaussian and is expressed as follows;
\begin{align}
\label{eq:joint}
y^{(t)}, \mathbf{y} \;\Big\vert\; \mu, \mathcal{X}, \ngbf{x}^{(t)}&\sim \bm{\mathcal{N}}\left(\mu\ngbf{1}_{n+1}, 
\begin{bmatrix}
\sigma_{\varepsilon}^2 + \sigma_{\gamma}^2 & \ngbf{c}^\top \\
\ngbf{c} & \gbf{\Sigma} + \sigma_{\gamma}^2\ngbf{I}
\end{bmatrix}
\right),
\end{align}
$$\sigma_{\varepsilon}^2=k(\ngbf{x}^{(t)}, \ngbf{x}^{(t)}), \quad \ngbf{c}_i = k(\ngbf{x}^{(t)}, \ngbf{x}^{(i)}), \quad \gbf{\Sigma}_{ij} = k(\ngbf{x}^{(i)}, \ngbf{x}^{(j)}),$$
where $\ngbf{1}_{n+1}$ denotes a column vector of length $n+1$ that contains only $1$'s. The homogeneous variance $\sigma_{\gamma}^2$ of the noise can be either determined by the user or estimated through maximum likelihood method. The \emph{posterior} distribution of $y^{(t)}$ can be calculated by marginalizing $\mu$ out and conditioning on the observed output variables $\mathbf{y}$~\cite{rasmussen2006gaussian}. Without any derivations, the posterior distribution for Ordinary Kriging is again Gaussian~\cite{ginsbourger2010kriging}:
\begin{equation}
y^{(t)} \;|\; \mathcal{X},\mathbf{y},\ngbf{x}^{(t)} \sim \bm{\mathcal{N}}\left(m(\ngbf{x}^{(t)}), s^2(\ngbf{x}^{(t)})\right) \label{eq:post} 
\end{equation}
where the posterior mean and variance are expressed as:
\begin{align}
m(\ngbf{x}^{(t)}) &= \hat{\mu} + \mathbf{c}^\top\left(\gbf{\Sigma} + \sigma_{\gamma}^2\ngbf{I}\right)^{-1}(\ngbf{y}-\hat{\mu}\ngbf{1}_n),\quad \hat{\mu} = \frac{\ngbf{1}_n^\top\left(\gbf{\Sigma} + \sigma_{\gamma}^2\ngbf{I}\right)^{-1}\ngbf{y}}{\ngbf{1}_n^\top\left(\gbf{\Sigma} + \sigma_{\gamma}^2\ngbf{I}\right)^{-1}\ngbf{1}_n}\label{eq:mean} \\
s^2(\ngbf{x}^{(t)}) &= \sigma_{\gamma}^2 + \sigma_{\varepsilon}^2 - \ngbf{c}^\top\left(\gbf{\Sigma} + \sigma_{\gamma}^2\ngbf{I}\right)^{-1}\ngbf{c} + \frac{(1-\ngbf{c}^\top\left(\gbf{\Sigma} + \sigma_{\gamma}^2\ngbf{I}\right)^{-1}\ngbf{1}_n)^2}{\ngbf{1}_n^\top\left(\gbf{\Sigma} + \sigma_{\gamma}^2\ngbf{I}\right)^{-1}\ngbf{1}_n} \label{eq:variance}
\end{align}
Note that the estimation of the trend, $\hat{\mu}$ is obtained by maximum a posteriori principle (MAP). The posterior mean function (Eq.~\ref{eq:mean}) is used as the predictor while the posterior variance (Eq.~\ref{eq:variance}) is the so-called \emph{Kriging variance} that measures the uncertainty of the prediction.

\section{Relevant Research}
\label{sec:Relevant}
Despite the theoretically sound development of the Kriging model, it suffers several issues when applied to large data sets. The major bottleneck is the high time complexity of the model fitting process: The inverse of the covariance matrix $\gbf{\Sigma}^{-1}$ needs to be computed for both the posterior mean and variance (Eq.~\ref{eq:mean} and~\ref{eq:variance}), which has roughly $O(n^3)$ time complexity\footnote{There are asymptotically faster algorithms for matrix inversion, e.g. Strasssen's $O(n^{2.807})$ and Stothers $O(n^{2.373})$, but their practical performance is worse than some methods with $o(n^3)$ time complexity.}. Moreover, when optimizing the hyper-parameters of the kernel function, the log likelihood function of those parameters is again calculated through $\gbf{\Sigma}^{-1}$, resulting in a $O(n^3)$ computational cost per each optimization iteration. Thus, for a large data set, such a high overhead in model fitting renders Kriging inapplicable in practice. Various attempts have been made to overcome the computational complexity issue of Kriging \cite{rasmussen2006gaussian}:
\begin{LaTeXdescription}
\item[Subset of Data] (SoD) \cite{lawrence2004gaussian} is a very simple approach in reducing complexity by taking $m < n$ data points, usually taken at random. Obviously this is a waste of information, but it might work well if a sufficient number of data points is available.

\item[Subset of Regressors] (SoR) \cite{silverman1985some} approximates Kriging by a linear combination of kernel functions on a set of basis points.
%For any input $\ngv{x_i}$, the function value $f_i$ is defined as:
%
%$$f_i = K_{i,u}\ngv{W_u},\ with\ \rho(\ngv{W_u})=\bm{\mathcal{N}}(0,K_{\ngv{u,u}}^{-1})$$
The basis points are linearly weighted to construct the predictor. The choice of the basis points \textit{does} influence the final outcome. As noted also in \cite{quinonero2005unifying}, there are only $m$ (number of basis points) degrees of freedom in the model because the model is degenerate (finite linear-in-the-parameters), which might be too restrictive.

\item[The Fully Independent Training Conditional] (FITC) \cite{naish2007generalized,snelson2005sparse}. Snelson and
 Ghahramani proposed what they called  Sparse  Gaussian  Processes  using  Pseudo-inputs. It uses a more sophisticated likelihood approximation with a richer covariance. It is a non-degenerate version of the \emph{SoR} algorithm. By providing a set of basis points (Pseudo inputs), the model is fitted and validated on the training data. As with \emph{SoR} the choice of basis points is a problem, this is usually either a subset of the training data or a uniform distribution over the input space.

\item[Fast Kriging with Gaussian Markov Random Fields]~\cite{Hartman2008} is an algorithm that uses an approximation of the covariance matrix with a sparse precision matrix. It uses \emph{Gaussian Markov Random Fields} (GMRF) on a reasonable dense grid to exploit the computational benefits of a Markov field while keeping the formula of Kriging weights. This method reduces the complexity for simple and ordinary Kriging, but might not always be efficient with universal Kriging. 

\item[Bayesian Committee Machines] (BCM) \cite{Tresp2000} is an algorithm similar to the one we propose, but developed from a completely different perspective. The basic motivation is to divide a huge training set into several relatively small subsets and then construct Kriging models on each subset. The benefit of this approach is that the training time on each subset is satisfactory and the training task can be easily parallelized. After training,  the prediction is made by weighted combination of estimations from all the Kriging models. BCM uses batch prediction to speed up the computation even further. However, BCM does not seem to correct for different hyper parameters per module, neither for badly fitted modules, which becomes a major problem when the number of modules increases.

Several other attempts have been made to divide the Kriging model in sub-models \cite{Chen2009,Nguyen-Tuong2009}, each solution for different domains. In \cite{Chen2009}, a \emph{Bagging}~\cite{Breiman1996} method is proposed to increase the robustness of the Kriging algorithm, rather than speeding up the algorithm's training time. In \cite{Nguyen-Tuong2009}, a partitioning method is introduced to separate the data points into local Kriging models and combine the different models using a distance metric.

\end{LaTeXdescription}

All of these approximation algorithms have their advantages and disadvantages and they are compared to our Cluster Kriging algorithms. 
The algorithms listed above can be divided into three categories: 
\begin{enumerate}
	\item methods that approximate the covariance matrix (using sparsity),
	\item methods that divide the training data into several clusters and build a model for each cluster,
	\item methods that take only a subset of the training data into account. 
\end{enumerate}
For the empirical study, three state of the art algorithms: \emph{SoD}, \emph{FITC} and \emph{BCM} are selected to compare with the proposed approaches in this paper,  as they seem to be the mostly used in their category.
%\footnote{There are also methods that use mathematical tricks to speed up the matrix inversion. These algorithms are not included in our comparison because each of these tricks can be used in combination with any of the other algorithms to reduce complexity.}

\section{Cluster Kriging}
\label{sec:Algorithm}
%ALGORITHMS

The main idea behind our proposed methodology, \emph{Cluster Kriging}, is to use a clustering algorithm to partition the data set into several clusters and build Kriging models on each cluster.
Loosely speaking, if the whole data set is partitioned into clusters of similar sizes, Cluster Kriging will reduce the time complexity by a factor of $k^2$ resulting in $k \cdot \left(\frac{n}{k}\right)^3$ (where $k$ is the number of clusters) if Kriging models are fitted sequentially. When exploiting $k$ CPU processes in parallel, the time complexity will be further reduced to $\left(\frac{n}{k}\right)^3$. In practice this means that if we take $k$ depending on $n$ our algorithm becomes quadratic in time, and using $k$ clusters it even reaches linear time complexity.
For the output value $y^{(t)}$ at an unobserved data point $\mathbf{x}^{(t)}$, each Kriging model provides a (local) prediction for $y^{(t)}$. To obtain a global prediction, it is proposed to either combine the predictions from all the Kriging models or select the most proper Kriging model for the prediction.  

There are many options for the data partitioning, e.g. K-means and Gaussian mixture models (GMM), and the Kriging model on clusters can also be combined in different manners. By varying the options in each step of the cluster Kriging, many algorithms can be generated. Four of them will be explained in the next section. In this section, the options in each step of the algorithms are introduced gradually.

\subsection{Clustering}

The first step in the Cluster Kriging methodology is the clustering of the input data $\mathcal{X}$ (and the output variables) into several smaller data sets. In general, the goal is to obtain a set $\mathcal{S}$ containing $k$ clusters on the input data set $\mathcal{X}$. 
\begin{equation}
\mathcal{S} = \{\mathcal{X}_1, \mathcal{X}_2, \ldots, \mathcal{X}_k\},
\quad \text{where } \bigcup_{i=1}^k\mathcal{X}_i = \mathcal{X}.
\end{equation}
In addition, the output values $\mathbf{y}$ are also grouped according to the clustering of $\mathcal{X}$: $\mathbf{y}=[\mathbf{y}_1^\top, \mathbf{y}_2^\top,\ldots, \mathbf{y}_{k}^\top]^\top$. The clustering can be done in many ways, with the most simple and feasible approach being random clustering. For our framework however we introduce three more sophisticated partitioning methods that are used in the experiments later on.

\subsubsection{Hard Clustering}
The hard clustering splits the data into $k$ smaller \emph{disjoint} data sets:
\begin{equation}
\bigcap_{i=1}^k\mathcal{X}_i = \emptyset \nonumber
\end{equation}
This can be achieved by various methods, for instance the K-means algorithm (Eq. \ref{eq:kmeans}).
K-means clustering minimizes the within-cluster sum of squares, that is expressed as:
\begin{equation}
\label{eq:kmeans}
\argmin_{\mathcal{S}} \sum_{i=1}^{k} \sum_{\mathbf{x} \in \mathcal{X}_i} ||\mathbf{x} - \gbf{\mu}^{(i)} ||^2,
\end{equation}
where $\gbf{\mu}^{(i)}$ is the centroid of cluster $i$ and is calculated as the mean of the points in $\mathcal{X}_i$. The minimization of the within-cluster sum of squares takes only $O(nkd)$ execution time.

\subsubsection{Soft Clustering}

Instead of using a hard clustering approach, a fuzzy clustering algorithm can be used to introduce slight overlap between the various smaller data sets, which might increase the final model accuracy. To incorporate fuzzy clustering, instead of directly applying cluster labels, the probabilities that a point belongs to a cluster are calculated (Eq. \ref{eq:fuzzyc}) and for each cluster $(n \cdot o)/k$ points with the highest membership values are assigned, where $o$ is a user defined setting that defines the overlap. $o$ is set between $1.0$ (no overlap) and $2.0$ (completely overlapping clusters). 

In principle, any fuzzy clustering algorithm can be used for the partitioning. In this paper the \emph{Fuzzy C-Means} (FCM) \cite{dunn1973fuzzy} clustering algorithm and the \emph{Gaussian Mixture Models} (GMM) \cite{reynolds2009gaussian} are used.
FCM is a clustering algorithm very similar to the well known \emph{K-means}. The algorithm differs from K-means in that it has additional membership coefficients and a fuzzifier. The membership coefficients of a given point give the degrees that this point belongs to each cluster. These coefficients are normalized so they sum up to one. The algorithm can be fitted on a given dataset and returns the coefficients for each data point to each cluster. The number of clusters is a user defined parameter. Fuzzy C-means optimizes the objective function given in Eq.~\ref{eq:fuzzyc} iteratively. In each iteration, the membership coefficients of each point being in the clusters are computed using Eq.~\ref{eq:m}. Subsequently, the centroid of each cluster $\gbf{\mu}^{(j)}$ is computed as the center of mass of all data points, taking the membership coefficients as weights. The objective of fuzzy C-means is to find a set of centroids that minimizes the following function:
\begin{equation}
\label{eq:fuzzyc}
\sum_{i=1}^{n} \sum_{j=1}^{k} w_{ij}^{m} ||\mathbf{x}^{(i)} - \gbf{\mu}^{(j)} ||^2,
\end{equation}
where $w_{ij}$ are the membership values (see Eq. \ref{eq:m}) and $m$ is the so-called fuzzifier (set to $2$ in this paper).  The fuzzifier determines the level of cluster fuzziness as follows:
\begin{equation}
\label{eq:m}
w_{ij}^{m} = \dfrac{1}{ \sum_{c=1}^{k} \biggl( \dfrac{|| \mathbf{x}^{(i)} - \gbf{\mu}^{(j)} ||}{|| \mathbf{x}^{(i)} - \gbf{\mu}^{(c)}||} \biggr)^{\frac{2}{m-1}} }
\end{equation}

The other fuzzy clustering procedure used is the Gaussian Mixture Models. GMM are used together with the \emph{expectation-maximization} (EM) algorithm for fitting the Gaussian models. The mixture models are fitted on the training data and later used in the weighted combination of the Kriging models by estimating cluster membership probabilities of the unseen data points. The advantage of this clustering technique is that it is fairly robust and that the number of clusters can be specified by the user. For the GMM method one could use the full covariance matrix whenever the dimensionality of the input data is small. However, when working with high dimensional data a diagonal covariance matrix can be used instead. The time complexity of GMM depends on the underlying EM algorithm. In each iteration EM, it takes $O(nk)$ operations to re-estimate the model parameters.

\subsubsection{Regression Tree Partitioning}
The third method used is the partitioning by use of a Regression Tree \cite{breiman1984classification} on the complete training set. The regression tree splits the dataset recursively at the best splitting point using the variance reduction criterion.
Each leaf node of the Regression Tree represents a cluster of data points. The number of leaves (or the number of records per leave) can be set by the user. By reducing the variance in each leaf node and therefore the variance in each dataset, the Kriging models can be fitted to the local datasets much better as will be presented later on. The time complexity of using a Regression Tree for the partitioning is $O(n)$, given that the depth of the tree or the number of leaf nodes is set by the user.
\begin{figure}[!htb]
      \centering
      \includegraphics[width=0.7\textwidth]{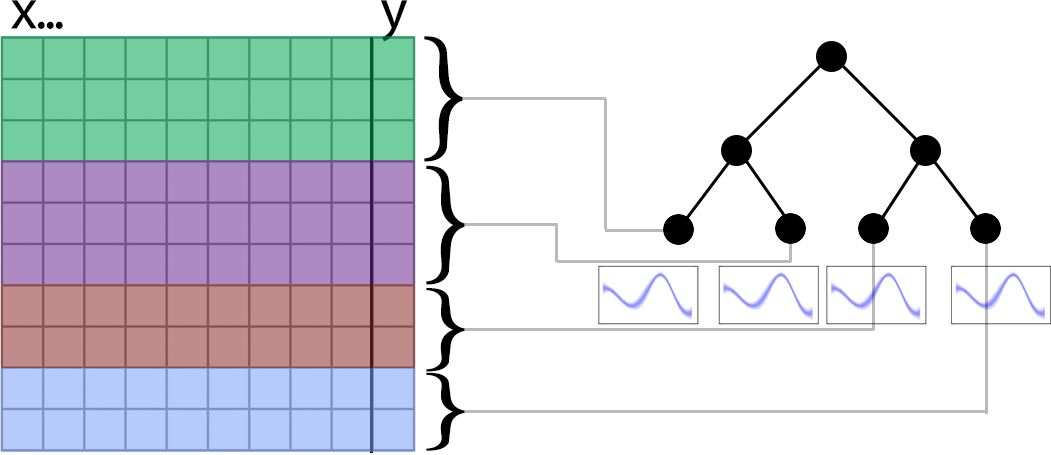}
      \caption{Visualisation of a Model Tree. The top node is the root and the bottom nodes are the leaves with attached models. Each record in the data (on the left) is assigned to a leaf node of the regression tree.\label{fig:modeltree}}
\end{figure}

\subsection{Modeling}
After partitioning the data set into several clusters, Kriging models are fitted on each of the smaller data sets. The Kriging algorithm is applied on each cluster individually, this way each model will be optimized on its own training set and will have different hyper-parameters. For simplicity we assume, in this paper, the kernel functions used on each cluster to be the same. As for the regression tree approach,  the data set, or more precisely the input space, is partitioned by the tree algorithm and, for each leaf node, a Kriging model is computed using the data belonging to this node (Fig.~\ref{fig:modeltree}). A similar technique is introduced in the context of combining linear regression models~\cite{wang1996induction,landwehr2005logistic,torgo1997functional}. In general, the predictive (posterior) distribution of the target variable $y^{(t)}$ on each cluster is:
\begin{equation}
\label{eq:post-cluster}
y^{(t)} \;|\; \mathcal{X}_l, \mathbf{y}_l, \ngbf{x}^{(t)} \sim \bm{\mathcal{N}}\left(m_l(\ngbf{x}^{(t)}), \sigma^2_l(\ngbf{x}^{(t)})\right), \quad l = 1,\ldots, k,
\end{equation}
where $m_l$ and $\sigma_l^2$ are specified again by Eq.~\ref{eq:mean} and~\ref{eq:variance} except that $\mathcal{X}, \mathbf{y}$ are replaced by $\mathcal{X}_l,\mathbf{y}_l$ here. Note that building the Kriging models can be easily parallelized, which gives an additional speedup to Cluster Kriging. Another benefit of building each model separately, is that each model has usually a much better local fit than a single global Kriging model would obtain.

\subsection{Prediction}

After training the various Kriging models, unseen data points need to be predicted. For this prediction, there are several options.
The first method which can be used is an optimally weighted combination of the Kriging models.

\subsubsection{Optimal weighting procedure}
\label{OWP}
When the input data set is separated by hard clustering methods, the Gaussian processes built on different clusters are independent from each other. In this sense, it is possible to construct a global Gaussian process model as the superposition of Gaussian processes from all the clusters. In addition, a weighting scheme is used to model how much ``trust'' should be put on the prediction from each cluster. The weighted superposition of all Gaussian processes is~\cite{van2015optimally}: 
\begin{equation}
\label{eq:post-final}
y^{(t)} \;|\; \mathcal{X}, \mathbf{y}, \mathbf{x}^{(t)} \sim \bm{\mathcal{N}}\left(\sum_{l=1}^{k}w_lm_l(\mathbf{x}^{(t)}), \sum_{l=1}^{k}w_l^2 \sigma_l^2(\mathbf{x}^{(t)})\right)
\end{equation}
The overall prediction and its variance depend on the weights used in the equation above. Intuitively, the optimal prediction is achieved when the variance of the estimation is minimal. To obtain such an optimal predictor, the overall Kriging variance should be optimized with respect to the weights, resulting in the following optimization task:
\begin{equation*}
\text{minimize:} \quad \sum_{l=1}^{k}w_l^2 \sigma_l^2(\mathbf{x}^{(t)}) \quad\text{subject to: } \sum_{l=1}^{k}w_l = 1, \quad w_l \geq 0, \quad l=1,\ldots,k.          
\end{equation*}
The optimal weights are obtained by solving the problem above (see the previous work of the authors~\cite{van2015optimally,van2016fuzzy} for details):
\begin{align}
w_l^* &= \frac{1/\sigma^{2}_l(\mathbf{x}^{(t)})}{\sum_{i=1}^k 1/\sigma^{2}_i(\mathbf{x}^{(t)})}. \label{eq:optimal}
\end{align} 
The optimal weights are then used to construct the optimal predictor, which is the inner product of the model predictions with the optimal weights.

\subsubsection{Membership Probabilities}
\label{MP}
For the GMM and other soft clustering approaches, the membership probabilities can be used for unseen records to define the weights for the combination of predictions. For each unseen record, the membership probabilities that this record belongs to the $k$ clusters are calculated and directly used as the weights in the weighted sum of predictions and variances given by the Kriging models:
\begin{align}
w_l &= \Pr(C=l \mid \mathcal{X}, \ngbf{x}^{(t)}), \quad \text{for } l = 1,\ldots,k \label{eq:membership}
\end{align} 
where $C$ is the cluster indicator variable ranging from $1$ to $k$. The rationale behind such a weighting scheme can be shown from the following derivation. In general, the goal here is to express the predictive distribution of variable $y^{(t)}$ that is the conditional density function on the whole data set $\mathcal{X}$, using the posterior densities from all clusters. By applying the total probability with respect to the cluster indicator variable $C$, such a density function $g$ can be written as~\cite{van2016fuzzy}:
\begin{align} 
g(y^{(t)} \mid \mathcal{X}, \ngbf{y}, \ngbf{x}^{(t)}) &= \sum_{l=1}^{k}g(y^{(t)}, C = l \mid \mathcal{X}, \ngbf{y}, \ngbf{x}^{(t)}) \nonumber\\
&\approx \sum_{l=1}^{k}g(y^{(t)} \mid \mathcal{X}_l, \ngbf{y}_l, \ngbf{x}^{(t)})\Pr(C=l \mid \mathcal{X}, \ngbf{x}^{(t)}) \label{EQ:12}
\end{align}
The independence assumption between Gaussian process models still holds approximately when the amount of the overlap between clusters is small. Thus, the density function $g(y^{(t)} \mid \mathcal{X}, \ngbf{y}, \ngbf{x}^{(t)})$ \emph{approximately} equals to Eq.~\ref{EQ:12}. The first term inside the sum in Eq.~\ref{EQ:12} is the predictive density function obtained from each cluster. The second term represents the probability of data point $\ngbf{x}^{(t)}$ belonging to a cluster, which is the weight in Eq.~\ref{eq:membership}. Consequently, the overall predictive density function is a \emph{mixture} of predictive distributions of all the Gaussian process models on clusters. To predict $y^{(t)}$, the expectation of the conditional density function $g(y^{(t)} \; | \; \mathcal{X}, \ngbf{y}, \ngbf{x}^{(t)} )$ is calculated:
\begin{align}
\E[y^{(t)} \mid \mathcal{X}, \ngbf{y}, \ngbf{x}^{(t)}] &= \sum_{l=1}^{k} \int_{-\infty}^{\infty} y^{(t)} \sum_{l=1}^{k}g(y^{(t)} \mid \mathcal{X}_l, \ngbf{y}_l, \ngbf{x}^{(t)})\Pr(C=l \mid \mathcal{X}, \ngbf{x}^{(t)}) \ud y^{(t)} \nonumber\\
&= \sum_{l=1}^{k}  \Pr(C=l \mid \mathcal{X}, \ngbf{x}^{(t)}) \E[y^t \; | \; \ngbf{X}_l, \ngbf{y}_l, \ngbf{x}^t] = \sum_{l=1}^{k} w_l m_l(\ngbf{x}^{(t)}) \label{EQ:14}
\end{align}
Note that $m_l(\ngbf{x}^t)$ is the mean function as shown in Eq.~\ref{eq:post-cluster}. Eq.~\ref{EQ:14} suggests that the overall prediction made on the whole data set can be expressed as a convex combination of the local predictions on each cluster of data, in which the combination weights are membership probabilities of GMM or similar clustering approaches. Furthermore, the variance of the prediction (expectation) above is derived as follows:
\begin{align}
&\Var[y^{(t)} \mid \mathcal{X}, \ngbf{y}, \ngbf{x}^{(t)}] \nonumber \\
&= \E[{y^{(t)}}^2 \mid \mathcal{X}, \ngbf{y}, \ngbf{x}^{(t)}] - \E[y^{(t)} \mid \mathcal{X}, \ngbf{y}, \ngbf{x}^{(t)}]^2 \nonumber \\
&=\sum_{l=1}^k w_l \left(\Var[y^{(t)} \mid \mathcal{X}_l, \ngbf{y}_l, \ngbf{x}^{(t)}] + \E[y^{(t)} \mid \mathcal{X}_l, \ngbf{y}_l, \ngbf{x}^{(t)}]^2\right) - \E[y^{(t)} \mid \mathcal{X}, \ngbf{y}, \ngbf{x}^{(t)}]^2  \nonumber \\
&=\sum_{l=1}^k w_l\left(\sigma_l^2(\mathbf{x}^{(t)}) + m_l(\ngbf{x}^{(t)})^2\right) - \left(\sum_{l=1}^{k} w_l m_l(\ngbf{x}^{(t)})\right)^2
\end{align}
Note that $\sigma_l^2(\mathbf{x}^{(t)})$ is again the Kriging variance at point $\mathbf{x}^{(t)}$ from cluster $l$.

\subsubsection{Single Model Prediction}
\label{SMP}
The last method which can be used to predict unseen data points is by using only one of the local Kriging models. First the partitioning method is used to predict which cluster the new data point belongs to, then the Kriging model trained using this particular cluster is used to predict the mean and variance at the new data point.
In case of the Regression Tree procedure, the targets are predicted from new unseen data points by first deciding which model needs to be used, using the Regression Tree. The target is then predicted using the specific Kriging model assigned to the leaf node (Figure \ref{fig:modeltree}). 
The main advantage of this method is that there is no combination of different predictions and only one of the local Kriging models needs to provide a prediction. This results in a significant speed-up for the prediction task.

\section{Flavors of Cluster Kriging}

Using the three stages and various components for each stage of the Cluster Kriging methodology, various algorithms can be implemented.
In this paper we asses four different flavors of Cluster Kriging:

\begin{LaTeXdescription}
\item{\textbf{Optimally Weighted Cluster Kriging}} (OWCK), which uses a hard (K-means) clustering technique to partition the data into $k$ clusters. Subsequently, a Kriging model is trained on each cluster and to predict unseen data points, the predictions and variances of each model are combined using the Optimal Weights Procedure (Section \ref{OWP}).
\item{\textbf{Optimally Weighted Fuzzy Cluster Kriging}} (OWFCK), which uses a soft clustering technique (Fuzzy C-Means) to partition the data into $k$  overlapping clusters and also uses the Optimal Weights Procedure combining the different predictions (Section \ref{OWP}). 
\item{\textbf{Gaussian Mixture Model Cluster Kriging}} (GMMCK), which uses Gaussian Mixture Models to partition the data into $k$ overlapping clusters and the trained Kriging models are weighted using the membership probabilities assigned on the unseen data by the Gaussian Mixture Model (Section \ref{MP}). 
\item{\textbf{Model Tree Cluster Kriging}} (MTCK), the proposed novel algorithm, uses a regression tree with a fixed amount of leaf nodes to partition the data in the objective space. A Kriging model is then trained on each partition defined by the tree's leaves. MTCK uses only one of the trained Kriging models per unseen record to predict (Section \ref{SMP}), depending on which leaf node the unseen record is assigned to.

% \begin{algorithm} [!ht]
% \caption{Model Tree Cluster Kriging}
% \label{MTCK}
% \begin{algorithmic}
% %
%   \REQUIRE {A data set $X_{train}$ with $n$ records, a target attribute $y$, an unseen dataset $X_{new}$ and the number of records per leaf $k$}
%   \STATE $\mathit{Tree} = \mathit{TrainRegressionTree}(X_{train}, k)$
%   \FORALL{$l \in \mathit{Tree.leaves}$}
%     \STATE $\mathit{Models}[l] = \mathit{GuassianProcess}(\mathit{l.getRecords}())$
%   \ENDFOR
% %
%   \FORALL{$x \in X_{new}$}
%     \STATE $l = \mathit{Tree.predict}(x)\mathit{.leaf}()$
%     \STATE $\mathit{Predictions}[x], \mathit{Variances}[x] = \mathit{model[l].predict}(x)$
%   \ENDFOR
%  \RETURN $\mathit{Predictions, Variances}$
% \end{algorithmic}
% \end{algorithm}

%Pseudo code of the MTCK algorithm is provided in Algorithm \ref{MTCK}.
First a decision tree regressor is constructed using the complete dataset. The tree is generated from the root node by recursively splitting the training data using the target variable and the variance reduction criterion. Once a node contains less than the minimum samples needed to split or the node contains only one record, the splitting stops and the node is called a leaf. To control the number of clusters, the user can set the maximum number of leaves or the minimum leaf size.
Next, each leaf node is assigned a unique index and each record belonging to the leaf is assigned to this index.
For each leaf, a Kriging model is computed using only those records assigned to this leaf. Each Kriging model is now able to predict a particular region defined by the Regression Tree.

For the prediction of the target for unseen records, the regression tree decides which Kriging model should be used. The final predicted mean and variance is provided by this Kriging model.

\end{LaTeXdescription}

\section{Experimental Setup and Results}

\label{sec:Res}

%results

A broad variety of experiments is executed to compare Optimally Weighted Cluster Kriging and its Fuzzy and Model Tree variants, to a wide set of other Kriging approximation algorithms. The algorithms included in the test are; \emph{Bayesian Committee Machines}, both with shared parameters (BCM sh.) and with individual parameters (BCM), \emph{Subset of Data} (SoD), \emph{Fully Independent Training Conditional} (FITC), \emph{Optimally Weighted Cluster Kriging} (OWCK) using K-means clustering, Fuzzy Cluster Kriging using Fuzzy C-means (OWFCK), Fuzzy Cluster Kriging with Gaussian Mixture Models (GMMCK) and, finally, Model Tree Cluster Kriging (MTCK).

The above algorithms are evaluated on three different data sets from the \emph{UCI machine learning repository}~\cite{Bache+Lichman:2013}:
\begin{itemize}
\item \emph{Concrete Strength} \cite{yeh1998modeling}, a data set with $1030$ records, $8$ attributes and one target attribute. The task is to predict the strength of concrete.
\item \emph{Combined Cycle Power Plant} (CCPP) \cite{Kaya2012}, a data set of $9568$ records, $3$ attributes and one target attribute. The target is the hourly electrical energy output and the task is to predict this target.
\item \emph{SARCOS} \cite{vijayakumar2005incremental}, a data set from \emph{gaussianprocess.org} with a training set of $44484$ records, $21$ attributes and $7$ target attributes. The task is to predict the joint torques of a anthropomorphic robot arm. All $21$ attributes are used as training data but only the $1^{st}$ target attribute is used as target. The dataset comes with a predefined test set of $4449$ records.
\end{itemize}
In addition, $8$ synthetic datasets with each $10.000$ records, $20$ attributes and one target attribute are used. The synthetic datasets are generated using benchmark functions from the Deap Python Package \cite{DEAP} and are often used in optimization. The functions are \emph{Ackley}, \emph{Schaffer}, \emph{Schwefel}, \emph{Rastrigin}, \emph{H1}, \emph{Rosenbrock}, \emph{Himmelblau} and \emph{Diffpow}.

\subsection{Hyper-Parameters}
\label{sec:hyper}
Each of the Kriging approximation algorithms has a hyper-parameter that can be tuned by the user to define the number of data points, clusters or inducing points, basically defining the trade-off between complexity and accuracy. For each of the algorithms a wide range of these hyper-parameters are used to see the effect and make a fair comparison between the different algorithms. The overlap for each of the Fuzzy algorithms is set to $10\%$, since from empirical experience we know that $10\%$ works well. Although higher percentages (above $10\%$) usually increase accuracy, the increase of accuracy is not significant and costs additional training time as well. For the Model Tree variant, the number of leaves is enforced by setting a minimum number of data points per leaf and an optional maximum number of leaves.
For the \emph{Concrete Strength} dataset and all synthetic datasets: FITC is set to a range of inducing points starting from $32$ and increasing in powers of $2$ to $512$.
SoD is set to the same range as FITC but for SoD this means the number of data points. BCM, both shared and non-shared versions and all \emph{Cluster Kriging} variants are set to a range from $2$ to $32$ clusters, increasing with powers of $2$.
For the \emph{Combined Cycle Power Plant} dataset: FITC is set to a range of inducing points starting from $64$ and increasing in powers of $2$ to $1024$.
\emph{SoD} is set to the a range from $256$ to $4092$ data points. BCM, both shared and non-shared versions and all \emph{Cluster Kriging} variants are set to a range from $4$ to $64$ clusters.

Finally, for the \emph{SARCOS} dataset, the range of FITC's inducing points stays the same as for the CCPP dataset, for SoD the range is from $512$ to $8184$ data points, and for all cluster based algorithms and the model tree variant, the range is set from $8$ to $128$ clusters.

\subsection{Quality Measurements} 

The quality of the experiments is estimated with the help of $5$-fold cross validation, except of the \emph{SARCOS} dataset, which uses its predefined test set.
The experiments are performed in a test framework similar to the framework proposed by \emph{Chalupka, K. et al.} \cite{Chalupka2012}, i.e. several quality measurements are used to evaluate the performance of each algorithm. The \emph{Coefficient of determination} $R^2$ score, \emph{Mean Standardized Log Loss} (MSLL) (see \cite{rasmussen2006gaussian} Chapter $8.1$) and the \emph{Standardized Mean Squared Error} (SMSE) are measured for each test run. 
The \emph{Mean Standardized Log Loss} is a measurement that takes both the predicted mean and the predicted variance into account. Penalizing wrong predictions that have a small predicted variance more than wrong predictions with a large variance.

$$\mathit{MSLL} = \left<\frac{1}{2} \cdot log(\pi\sigma_i + (y_i-\hat{y}_i)^2 / \sigma_i) - \mathit{triv}\right> $$
Where $\sigma^t$ is the predicted variance for record $\ngbf{x}_i$ and $\hat{y}_i$ the predicted mean.
With $\mathit{triv}$ the trivial score simulating a predictor that predicts the overall mean and standard deviation:
$$\mathit{triv} = \frac{1}{2} \cdot log(\pi\sigma_y + (y_i-\bar{y})^2 / \sigma_y) $$

For MSLL and SMSE lower scores are better, for $R^2$, $1.0$ is the best possible score meaning a perfect fit and everything lower is worse.

\subsection{Results}

\begin{table*}[!h]
\renewcommand{\arraystretch}{1.3}
\caption{Average $R^2$ score per dataset for each algorithm}
\label{table:r2all}
\centering
\begin{tabular}{l|c c c c c c c c c c c}
 \hline
\bfseries {Dataset} & {SOD} & {OWCK} & {GMMCK} & {OWFCK} & {FITC} & {BCM} & {BCM sh.} & {MTCK} \\
\hline
{ concrete } &  0.784 & 0.826 & 0.839 & 0.696 & 0.675 & -81.888 & -242.459 & \bfseries 0.851 & \\
{ CCPP } &  0.948 & 0.937 &\bfseries  0.968 & 0.916 & 0.890 & 0.220 & -24.602 &  \bfseries 0.968 & \\
{ sarcos } &  0.964 & 0.894 & 0.996 & 0.570 & 0.941 & -627.280 & 0.448 & \bfseries 0.999 & \\
{ ackley } &  0.952 & 0.957 & 0.951 & 0.954 & 0.260 & 0.921 & -0.039 & \bfseries 0.981 & \\
{ schaffer } &  0.321 & 0.388 & 0.369 & 0.406 & 0.208 & 0.452 & -0.050 & \bfseries 0.672 & \\
{ schwefel } &  0.990 & 0.973 & 0.977 & 0.947 & 0.006 & 0.969 & -0.043 & \bfseries 0.999 & \\
{ rast } &  0.973 & 0.947 & 0.948 & 0.932 & 0.322 & 0.914 & -0.043 &\bfseries  0.998 & \\
{ h1 } &  0.676 & -0.082 & 0.527 & -1.125 & 0.165 & 0.657 & -0.046 & \bfseries 0.977 & \\
{ rosenbrock } &  0.999 & 0.997 & 0.997 & 0.981 & 0.000 & 0.994 & -0.050 &\bfseries  1.000 & \\
{ himmelblau } &  0.997 & 0.995 & 0.995 & 0.981 & 0.291 & 0.994 & -0.044 & \bfseries 1.000 & \\
{ diffpow } &  0.995 & 0.991 & 0.991 & 0.975 & 0.001 & -0.001 & -0.001 &\bfseries  1.000 & \\
\hline
\end{tabular}
\end{table*}

\begin{table*}[!h]
\renewcommand{\arraystretch}{1.3}
\caption{Average MSLL score per dataset for each algorithm}
\label{table:msslall}
\centering
\begin{tabular}{l|c c c c c c c c c c c}
 \hline
\bfseries{Dataset} & {SOD} & {OWCK} & {GMMCK} & {OWFCK} & {FITC} & {BCM} & {BCM sh.} & {MTCK} \\
\hline
{ concrete } &  -0.837 & -0.946 & -1.100 & -0.692 & -0.629 & 18.590 & 68.013 & \bfseries -1.140 & \\
{ CCPP } &  -0.089 & -1.438 & \bfseries -1.525 & -1.109 & -1.165 & 7.826 & 69.346 & -1.193 & \\
{ sarcos } &  -1.926 & -1.371 & -3.147 & -0.302 & -1.463 & 780.090 & 507.721 & \bfseries -3.429 & \\
{ ackley } &  -1.622 & -1.516 & -1.517 & -1.462 & -0.104 & 7.352 & 13.010 & \bfseries -2.012 & \\
{ schaffer } &  0.477 & -0.073 & 0.081 & -0.091 & -0.107 & 16.872 & 11.707 & \bfseries -0.514 & \\
{ schwefel } &  -2.554 & -2.013 & -2.162 & -1.944 & -0.002 & -0.144 & 12.034 & \bfseries -3.278 & \\
{ rast } &  -2.179 & -1.686 & -1.807 & -1.642 & -0.193 & 4.554 & 11.590 & \bfseries -2.901 & \\
{ h1 } &  -0.766 & -0.276 & -0.540 & -0.060 & -0.059 & 9.018 & 17.393 & \bfseries -1.967 & \\
{ rosenbrock } &  -3.479 & -2.915 & -3.074 & -2.738 & high* & 0.612 & 18.575 & \bfseries -4.054 & \\
{ himmelblau } &  -3.204 & -2.646 & -2.790 & -2.553 & -0.193 & -1.422 & 12.826 & \bfseries -3.739 & \\
{ diffpow } &  -3.020 & -2.548 & -2.666 & -2.438 & high* & high* & high* & \bfseries -3.744 & \\
\hline
\end{tabular}
\end{table*}

\begin{table*}[!h]
\renewcommand{\arraystretch}{1.3}
\caption{Average SMSE score per dataset for each algorithm}
\label{table:smseall}
\centering
\begin{tabular}{l|c c c c c c c c c c c}
 \hline
\bfseries{Dataset} & {SOD} & {OWCK} & {GMM-CK} & {FCM-CK} & {FITC} & {BCM} & {BCM sh.} & {MTCK} \\
\hline
{ concrete } &  0.216 & 0.174 & 0.161 & 0.304 & 0.325 & 82.888 & 243.459 & \bfseries 0.149 & \\
{ CCPP } &  0.052 & 0.063 & \bfseries 0.032 & 0.084 & 0.110 & 0.780 & 25.602 & \bfseries 0.032 & \\
{ sarcos } &  0.036 & 0.106 & 0.004 & 0.430 & 0.059 & 628.280 & 0.552 & \bfseries 0.001 & \\
{ ackley } &  0.048 & 0.043 & 0.049 & 0.046 & 0.740 & 0.079 & 1.039 &\bfseries 0.019 & \\
{ schaffer } &  0.679 & 0.612 & 0.631 & 0.594 & 0.792 & 0.548 & 1.050 &\bfseries 0.328 & \\
{ schwefel } &  0.010 & 0.027 & 0.023 & 0.053 & 0.994 & 0.031 & 1.043 &\bfseries 0.001 & \\
{ rast } &  0.027 & 0.053 & 0.052 & 0.068 & 0.678 & 0.086 & 1.043 & \bfseries0.002 & \\
{ h1 } &  0.324 & 1.082 & 0.473 & 2.125 & 0.835 & 0.343 & 1.046 & \bfseries0.023 & \\
{ rosenbrock } &  0.001 & 0.003 & 0.003 & 0.019 & 1.000 & 0.006 & 1.050 & \bfseries0.000 & \\
{ himmelblau } &  0.003 & 0.005 & 0.005 & 0.019 & 0.709 & 0.006 & 1.044 & \bfseries0.000 & \\
{ diffpow } &  0.005 & 0.009 & 0.009 & 0.025 & 0.999 & 1.001 & 1.001 & \bfseries0.000 & \\
\hline
\end{tabular}
\vspace{-1cm}
\end{table*}

\begin{figure*}[!b]
    \centering
    \begin{subfigure}[t]{0.60\textwidth}
        \centering
        \includegraphics[width=\textwidth]{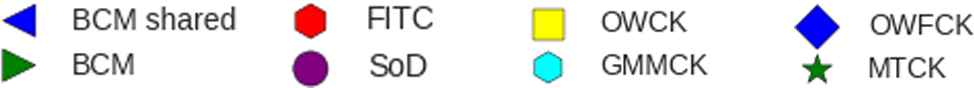}
    \end{subfigure}

    \begin{subfigure}[t]{0.47\textwidth}
        \centering
        \includegraphics[width=\textwidth]{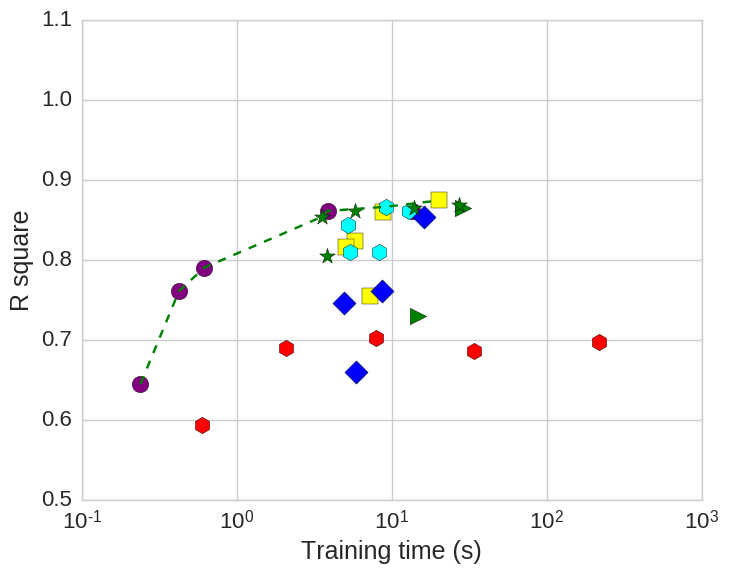}
        \caption{Concrete}
    \end{subfigure}
    ~ 
    \begin{subfigure}[t]{0.47\textwidth}
        \centering
        \includegraphics[width=\textwidth]{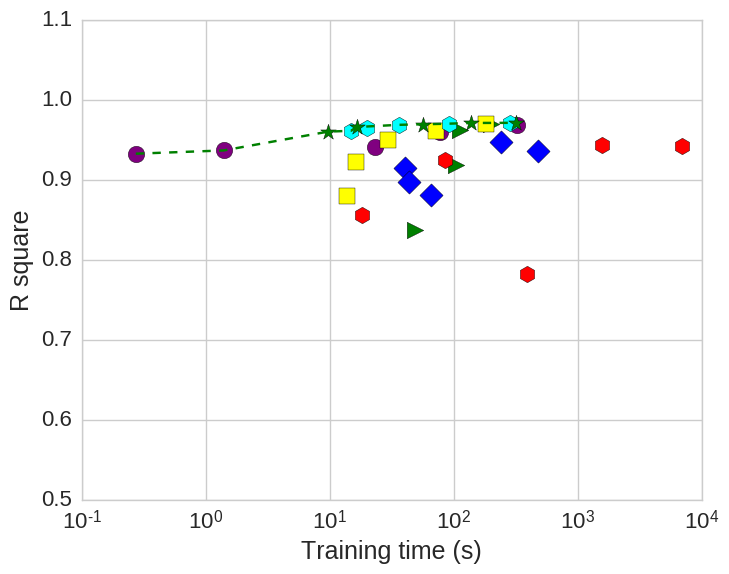}
        \caption{CCPP}
    \end{subfigure}
    ~
    \begin{subfigure}[t]{0.47\textwidth}
        \centering
        \includegraphics[width=\textwidth]{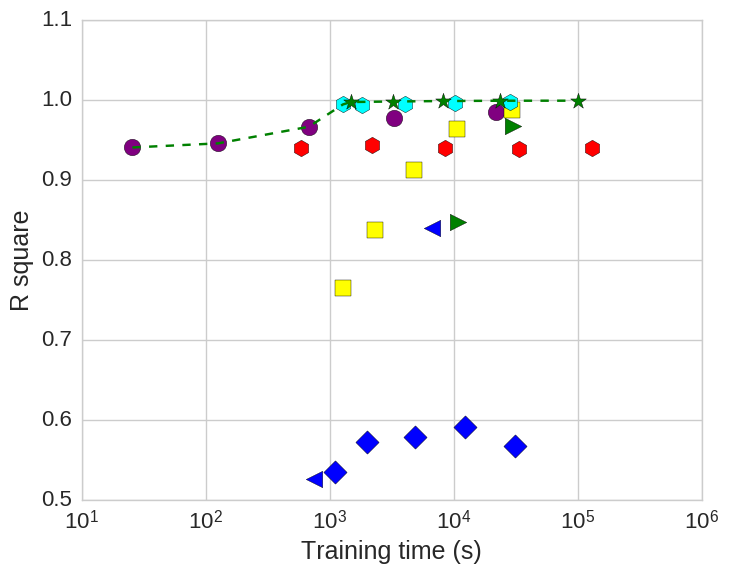}
        \caption{Sarcos}
    \end{subfigure}
    ~
    \begin{subfigure}[t]{0.47\textwidth}
        \centering
        \includegraphics[width=\textwidth]{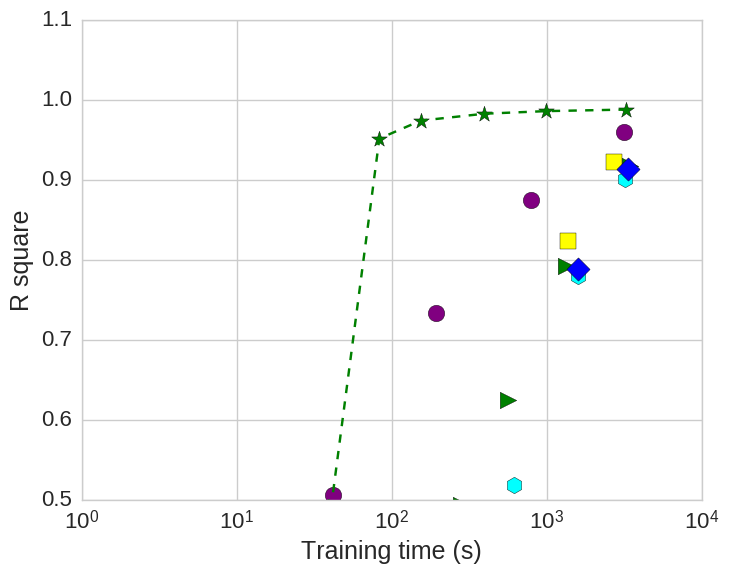}
        \caption{H1}
    \end{subfigure}

 \caption{Quality measurements of each algorithm with the hyper-parameters increasing in sample sizes for FITC and SoD, and decreasing in cluster sizes for the cluster based algorithms as explained in Section \ref{sec:hyper}. The results are shown for the Concrete, CCPP and Sarcos datasets and the gerenated dataset for the H1 function. The training time is given on the $x$ axis and the $R^2$ score on the y axis. The dashed green line indicates the non-dominated set. \label{fig:allreal}}
\vspace{-1cm}
\end{figure*}

The results of experiments on the real world datasets \emph{Concrete Strength}, \emph{CCPP} and \emph{SARCOS} are shown in Figure \ref{fig:allreal}. The results are shown with both objectives, time and accuracy ($x$ and $y$ axis respectively) in mind to show the trade-off and to show that some algorithms are performing better in both objectives. The $R^2$ scores of each dataset per algorithm, averaged over all folds, are shown in Table \ref{table:r2all}. The MSLL scores are provided in Table \ref{table:msslall} and the SMSE scores in Table \ref{table:smseall}. The best results for each dataset are shown in bold face.

\subsection{Parameter Setting Recommendations}

To use the Cluster Kriging algorithms, the minimum cluster size or the number of clusters has to be set as a user defined parameter. It is recommended to set this parameter in such a way that each individual cluster contains between $100$ and $1000$ records. $1000$ records is still computationally tractable by Kriging in terms of execution time and $100$ records is in most cases still doable in terms of fitting the Kriging model. Selecting smaller cluster sizes is likely to result in poorly fitted models and selecting cluster sizes larger than $1000$ will in most cases not increase accuracy but will only increase execution time. These recommendations are purely based on empirical observations and depend highly on the dataset one is working with. For MTCK smaller cluster sizes are usually still fine because of the low variance in the records per leaf due to the splitting criterion of the Regression Tree.

%SOD 0.985031621725 & -2.35871638429 & 0.0149683782747 \\
%FITC 0.944834532574 & -1.65303974123 & 0.0551654674258 \\
%BCM 0.967949648722 & 1.90276832257 & 0.0320503512783 \\
%BCM shared 0.840209358742 & -0.645788225247 & 0.159790641258 \\
%KM_indices 0.987610062829 & -2.14739971359 & 0.0123899371709 \\
%GMM_indices 0.998363616138 & -3.45470426223 & 0.00163638386245 \\
%F_indices 0.591976658419 & -0.424409292265 & 0.408023341581 \\
%FLAME 0.937418699923 & -1.78505904737 & 0.0625813000774 \\

\section{Conclusions and Further Research}

\label{sec:Con}
%conclusions
A  novel Kriging approximation methodology, Cluster Kriging, is proposed, using a combination of smaller Kriging models trained on partitions of the data set. Four different algorithms using this methodology are proposed and explained in detail and a broad comparison between the novel algorithms and other state of the art Kriging approximation algorithms is done. The results of the experiments (as given in Section \ref{sec:Res}) clearly show that for each data set, the \emph{Gaussian Mixture Models Cluster Kriging} (GMM CK) and the Model Tree Cluster Kriging  (MTCK) outperform the other algorithms in all measurements. It can also be observed that the \emph{Bayesian Committee Machine} algorithms, both with shared parameters and with individual parameters, are very unstable when the number of clusters is above $8$. This is most likely due to poor recombination of models with different hyper-parameters and the chance of poor fitting of one of the clusters. In terms of training time, \emph{Subset of Data} is much faster than any of the other algorithms, though it pays for this complexity reduction by a decrease in accuracy. Both for SoD and FITC, the training time increases faster than the training time of the cluster based algorithms. It is shown that the membership probabilities of the Gaussian Mixture Model can be used as weights in the combination of the various Kriging models' predictions.  It is also shown that a Model Tree of Kriging models works very well in high dimensional problems and requires less prediction time due to the fact that only one Kriging model per unseen data point is used for prediction.

For future research it would be interesting to automatically determine cluster sizes for the different algorithms and optimize the nugget parameter of the Kriging models. The nugget plays an important role in the fitting of the Kriging model as it determines the amount of marginalization.

\section*{Acknowledgment}
The authors acknowledge support by NWO (Netherlands Organisation for Scientific Research) PROMIMOOC project (project number: 650.002.001).

\FloatBarrier

\bibliography{main}

\end{document}